\documentclass[letterpaper, 10 pt, conference]{ieeeconf}  

\IEEEoverridecommandlockouts                              

\usepackage{graphics} 
\usepackage{amsmath} 
\usepackage{amssymb}  
\usepackage{algorithm}
\usepackage{hyperref} 
\usepackage{graphicx}
\usepackage{cleveref}
\usepackage{pifont}
\usepackage{multirow}
\usepackage{booktabs}
\usepackage{wrapfig}
\usepackage{caption}
\usepackage[dvipsnames, table]{xcolor}  
\usepackage{pifont}
\usepackage{mwe}
\usepackage{xspace}
\usepackage{footmisc} 
\usepackage{listings}
\usepackage{lipsum}
\usepackage{adjustbox} 
\usepackage{hyperref}

\usepackage[normalem]{ulem}
\useunder{\uline}{\ul}{}

\newcommand{\ourmethod}{\emph{SysNav}\xspace}
\newcommand{\ourmethodplain}{{SysNav}\xspace}
\definecolor{myred}{RGB}{231,76,60}
\definecolor{myblue}{RGB}{52,152,219}
\definecolor{myorange}{RGB}{243,156,18}
\definecolor{cvprblue}{rgb}{0.21,0.49,0.74}
\hypersetup{
    colorlinks=true,
    linkcolor=cvprblue,
    citecolor=cvprblue,
    urlcolor=cvprblue
}

\crefname{equation}{Eq.}{Eqs.} \crefname{section}{Sec.}{Secs.} \crefname{subsection}{Sec.}{Secs.}
\crefname{figure}{Fig.}{Figs.} \crefname{table}{Tab.}{Tabs.} \crefname{algorithm}{Alg.}{Algs.}
\crefname{theorem}{Thm.}{Thms.} \crefname{lemma}{Lem.}{Lems.} \crefname{corollary}{Cor.}{Cors.}
\crefname{proposition}{Prop.}{Props.} \crefname{definition}{Def.}{Defs.}
\crefname{remark}{Rem.}{Rems.} \crefname{example}{Ex.}{Exs.} \crefname{appendix}{App.}{Apps.}
\crefname{line}{Line}{Lines} \crefname{footnote}{Fn.}{Fns.} \crefname{enumi}{Item}{Items}

\title{\LARGE \bf
SysNav: Multi-Level Systematic Cooperation Enables\\Real-World, Cross-Embodiment Object Navigation
}

\author{
	\textbf{Haokun Zhu}$^{1}$,  
	\textbf{Zongtai Li}$^{1}$, 
	\textbf{Zihan Liu}$^{1,2}$, 
	\textbf{Kevin Guo}$^{1}$, 
	\textbf{Zhengzhi Lin}$^{1}$, \\
	\textbf{Yuxin Cai}$^{1,3}$,
	\textbf{Guofei Chen}$^{1}$,
	\textbf{Chen Lv}$^{3}$, 
	\textbf{Wenshan Wang}$^{1}$,
	\textbf{Jean Oh}$^{1}$,
	\textbf{Ji Zhang}$^{1}$ \\
	$^{1}$Carnegie Mellon University \quad
	$^{2}$New York University \quad
	$^{3}$Nanyang Technological University \\
}

\begin{document}


\twocolumn[{%
\renewcommand\twocolumn[1][]{#1}%
\maketitle
\begin{center}
    \centering \small
    \captionsetup{type=figure}
    \includegraphics[width=0.98\linewidth]{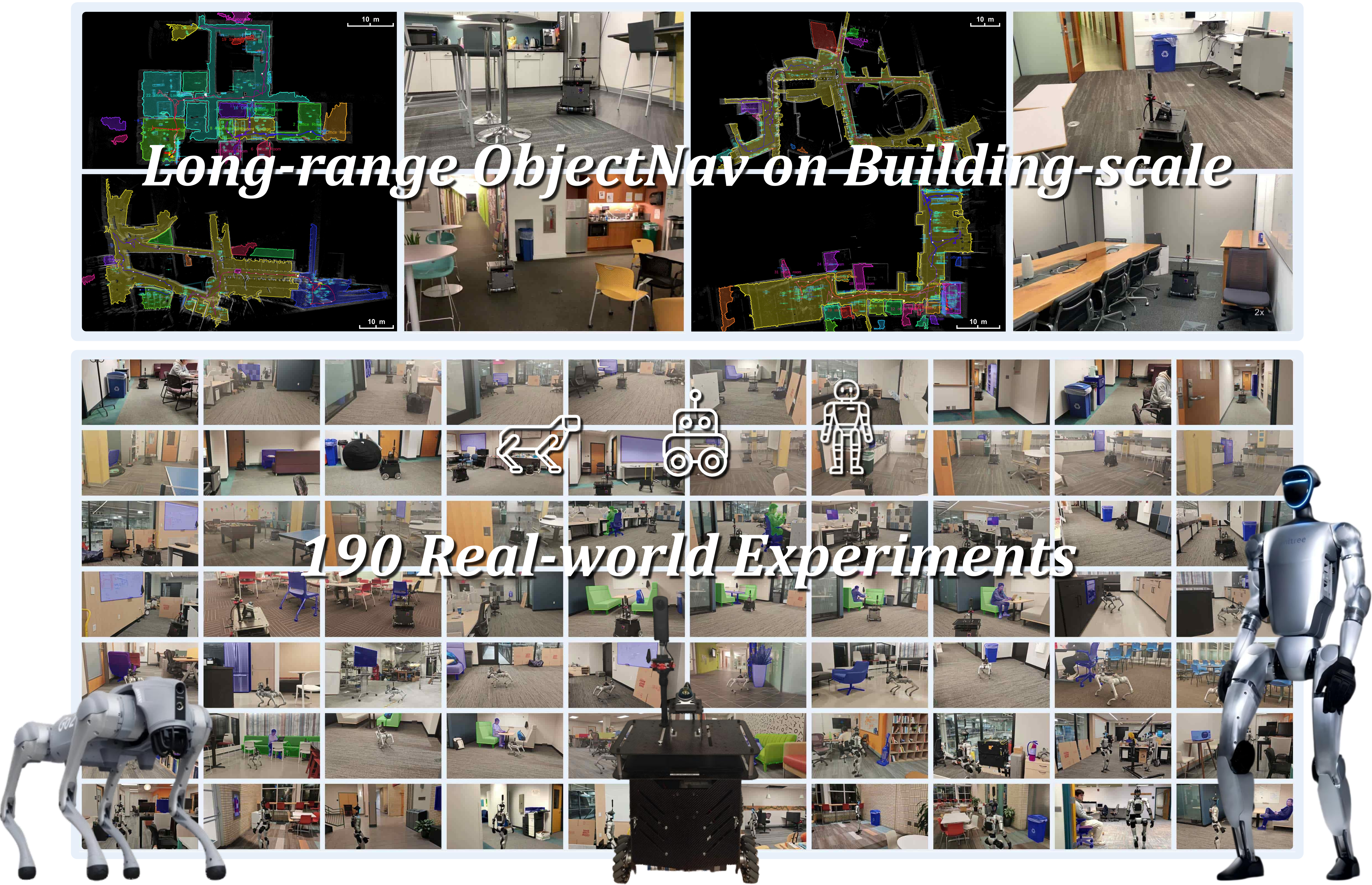}
    \vspace{-3pt}
    \captionof{figure}{{\ourmethodplain achieves object navigation in diverse real-world environments and generalizes across multiple embodiments. To our knowledge, it is the first system to reliably and efficiently complete object navigation at building-scale.}}\vspace{-5pt}
    \label{fig:teaser}
\end{center}%
}]

\begin{abstract}
Object navigation (ObjectNav) in real-world environments is a complex problem that requires simultaneously addressing multiple challenges, including complex spatial structure, long-horizon planning and semantic understanding.
Recent advances in Vision-Language Models (VLMs) offer promising capabilities for semantic understanding, yet effectively integrating them into real-world navigation systems remains a non-trivial challenge.
In this work, we formulate real-world ObjectNav as a system-level problem and introduce \ourmethod, a three-level ObjectNav system designed for real-world cross-embodiment deployment.
\ourmethodplain decouples semantic reasoning, navigation planning and motion control to ensure robustness and generalizability.
At the high-level, we summarize the environment into a structured scene representation and leverage VLMs to provide semantic-grounded navigation guidance.
At the mid-level, we introduce a hierarchical room-based navigation strategy that reserves VLM guidance for room-level decisions, which effectively utilizes its reasoning ability while ensuring system efficiency.
At the low-level, planned waypoints are executed through different embodiment-specific motion control modules.
We deploy our system on three embodiments, a custom-built wheeled robot, the Unitree Go2 quadruped and the Unitree G1 humanoid, and conduct 190 real-world experiments. Our system achieves substantial improvements in both success rate and navigation efficiency. To the best of our knowledge, \ourmethodplain is the first system capable of reliably and efficiently completing building-scale long-range object navigation in complex real-world environments. Furthermore, extensive experiments on four simulation benchmarks demonstrate state-of-the-art performance.
Project page is available at \href{https://cmu-vln.github.io/}{\textit{https://cmu-vln.github.io/}}.
\end{abstract}


\vspace{-3pt}
\section{Introduction}
\label{sec:introduction}
Object Navigation (ObjectNav) is a fundamental problem in robotics, where an agent must autonomously locate and reach an instance of a target object category in an unknown environment as soon as possible. Successfully accomplishing this task requires not only robust visual perception, but also long-horizon decision making and semantic understanding. While recent years have witnessed significant progress in ObjectNav within simulation, deploying real-world ObjectNav systems that are both stable and efficient, particularly across different embodiments, remains a challenging problem.

A primary challenge of real-world ObjectNav is that it is inherently a complex, multifaceted problem composed of multiple sub-challenges, including complex spatial structure, long-horizon planning and semantic understanding.
Despite this complexity, most existing approaches formulate ObjectNav as a single-policy learning problem, typically relying on end-to-end models that directly map raw sensor observations to discrete actions.
However, a single end-to-end policy struggles to simultaneously address these diverse challenges, and the scarcity of real-robot training data further exacerbates the difficulty of applying such approaches in real-world settings.
Therefore, we argue that real-world ObjectNav should be treated as a system-level problem rather than a single-policy learning task, requiring a fully integrated multi-level system that explicitly decouples different sub-challenges and designs specialized modules to address them.

Another critical challenge lies in how to effectively organize semantic information and leverage the reasoning capabilities of VLMs for real-world navigation. Although VLMs exhibit strong commonsense reasoning abilities, many existing approaches over-rely on them for exploration and decision-making. While such strategies may perform adequately in clean simulation environments, they often perform poorly in complex real-world settings. This is because ObjectNav requires precise spatial grounding and long-term spatial consistency, capabilities that current VLMs do not inherently possess.
We argue that a successful ObjectNav system should utilize VLM reasoning abilities at the appropriate spatial level, rather than overextending its responsibilities beyond their effective capacity. Meanwhile, the system should organize environmental information in a structured form to provide meaningful contextual information, allowing VLMs to effectively leverage their reasoning strengths.

To address these challenges, we propose \ourmethod, a three-level ObjectNav system designed for real-world deployment and cross-embodiment generalization. \ourmethodplain decouples semantic reasoning, navigation planning and motion control, allowing each component to focus on its respective strengths.
The high-level module organizes environmental information into a structured scene representation and the VLM performs high-level semantic reasoning over this representation, providing semantic-grounded navigation guidance.
Guided by the high-level semantic reasoning, the mid-level module plans navigation strategies, exploration paths and waypoints, while the low-level module executes these waypoints via embodiment-specific motion control.
At the mid-level, we further introduce a hierarchical navigation strategy that treats rooms as the minimal decision-making units for the VLM.
Indoor environments naturally decompose into functionally distinct rooms (e.g., kitchens, bedrooms), where target objects exhibit strong semantic priors (e.g., refrigerators in kitchens), while the spatial scale of individual rooms allows efficient coverage using classical exploration methods.
Moreover, given VLMs' limited 3D spatial understanding~\cite{chen2024spatialvlm, qi2025gpt4scene}, involving VLMs in fine-grained, in-room exploration fails to leverage their semantic strengths and may reduce efficiency.
Based on this insight, we restrict VLM-based reasoning to room-level decisions and employ efficient classical exploration within rooms.
Supported by the structured scene representation, the VLM selects the next room to explore, enabling efficient and robust global navigation.

We have successfully deployed the proposed system across three robot embodiments, a wheeled robot, the Unitree Go2 quadruped, and the Unitree G1 humanoid, and conducted 190 real-world experiments. Our system achieves a {4-5$\times$} improvement in navigation efficiency over existing ObjectNav baselines and, to the best of our knowledge, is the first system capable of reliably and efficiently completing object navigation at building-scale. In addition, we evaluate our approach on four widely used simulation benchmarks, HM3D-v1, HM3D-v2, MP3D, and HM3D-OVON, where it achieves state-of-the-art performance across all benchmarks.
\begin{figure*}[t]
    \centering
    \includegraphics[width=1.0\linewidth]{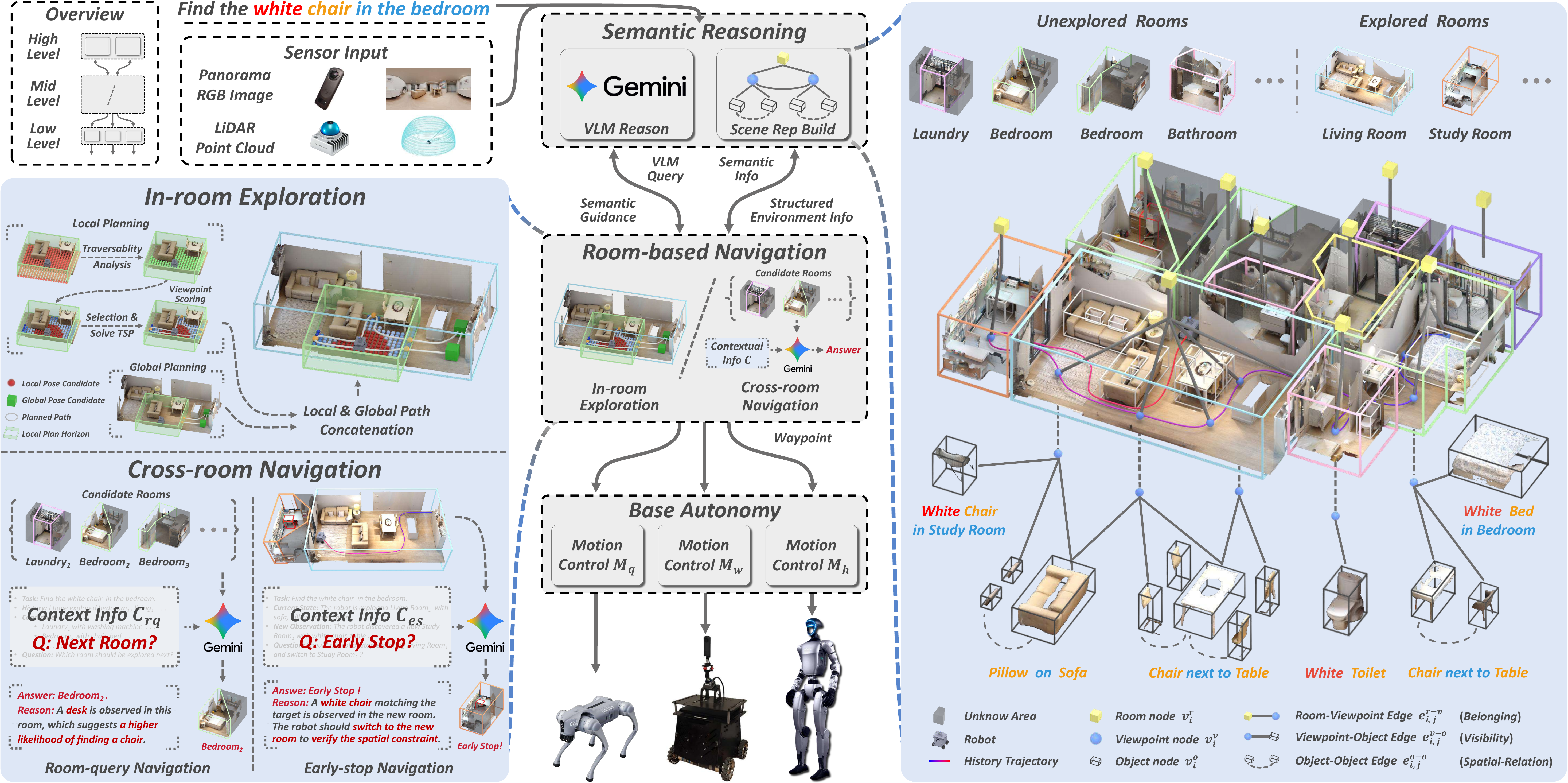}
    \caption{\textbf{Overview of the proposed \ourmethodplain.} The high-level {\ul \textit{Semantic Reasoning}} module organizes environmental information into a structured scene representation (\cref{sec:method_scenerepresentation}) and leverages VLM reasoning to provide semantic-grounded navigation guidance (\cref{sec:method_vlmreasoning}). The mid-level {\ul \textit{Room-based Navigation}} module performs hierarchical navigation with in-room exploration (\cref{sec:method_inroom}) and cross-room navigation (\cref{sec:method_crossroom}). The low-level {\ul \textit{Base Autonomy}} module executes planned waypoints through embodiment-specific motion control (\cref{sec:method_lowlevel}).}
    \vspace{-15pt}
    \label{fig:piepline}
\end{figure*}

\vspace{-8pt}
\section{Related Works}
\label{sec:related_works}

\subsection{Object Navigation}
\label{sec:related_works_object_navigation}
Existing ObjectNav methods fall into two categories: end-to-end learning and modular approaches. End-to-end methods~\cite{wijmans2019dd, ye2021auxiliary, ramrakhya2022habitat, ramrakhya2023pirlnav, zhang2024uni} use reinforcement learning or VLM fine-tuning to directly map observations to discrete actions.
In contrast, modular methods~\cite{chaplot2020object, ramakrishnan2022poni, zhang20233dawareobjectgoalnavigation, Yu_2023, zhou2023escexplorationsoftcommonsense, yokoyama2024vlfm, yin2024sg, zhu2025move, zhang2025apexnav} decompose navigation into steps such as mapping, planning, and action execution, and often leveraging semantic maps for more interpretable and scalable navigation behavior.
However, most existing methods are developed in simplified simulation settings and treat ObjectNav primarily as a policy learning problem, limiting their ability to handle real-world complexity and real-time constraints. To address this gap, we propose a comprehensive three-level ObjectNav system designed for complex real-world environments and cross-embodiment deployment.

\subsection{VLM-Guided Navigation}
\label{sec:related_works_vlm_guided_navigation}
With internet-scale training data, Vision-Language Models (VLMs)~\cite{li2022blipbootstrappinglanguageimagepretraining, radford2021learning, wu2024gpt4visionhumanalignedevaluatortextto3d, team2023gemini} have shown strong commen-sense reasoning abilities and have been widely applied in Object Navigation tasks to guide the decision-making process. For example, InstructNav~\cite{longinstructnav} leverages multi-sourced value maps to model key navigation elements. ApexNav~\cite{zhang2025apexnav} employs VLMs to score each frontier for exploration planning. However, due to VLMs' limited 3D spatial understanding ability~\cite{zhang2024vision, chen2024spatialvlm, qi2025gpt4scene}, over-reliance on VLMs for navigation can lead to inefficient behavior, such as frequent backtracking. To address this, we propose to treat rooms as the minimal decision-making unit for VLMs to provide semantic-grounded navigation guidance.

\subsection{Scene Representation for Indoor Navigation}
\label{sec:related_works_scene_representation}
Scene representation~\cite{hughes2022hydra} transforms raw observations into structured information for navigation. Frontier-based methods~\cite{ramakrishnan2022poni, chen2023not, gadre2023cows, gervet2023navigating, zhang2025apexnav} record frontiers on grid maps and incorporate semantic cues for exploration.
In contrast, graph-based methods represent environments as scene graphs to support higher-level reasoning. Prior works~\cite{yin2024sg, loo2024open} use scene graphs to summarize semantic information and let VLMs to select frontier locations. Others~\cite{an2024etpnav,zhu2025strive, wu2024voronav} explicitly construct viewpoints in the scene graph, enabling VLMs to reason over the graph and choose among viewpoints to guide navigation.
While effective, existing scene representation in ObjectNav often lack spatial structure and contextual information for high-level reasoning.
We propose a three-layer scene representation that models multi-granularity nodes and their spatial relationships, providing structured context for VLM reasoning.
\section{Problem Formulation}
\label{sec:problem_formulation}
The robot is initialized in an unknown indoor environment $\mathcal{M}$ containing multiple objects $\mathcal{O}$. At each time step, it receives a multimodal observation of the environment:
$o_t = \{ I_t^{360}, P_t^{lidar} \}$, where $I_t^{360} \in \mathbb{R}^{H \times W \times 3}$ denotes a 360$^\circ$ RGB panoramic image captured at the robot's current pose, and $P_t^{lidar} \subset \mathbb{R}^3$ denotes the LiDAR point cloud. 
The robot executes actions $a_t = (\vec{v}_t, \omega_t)$, parameterized by linear and angular velocities, from a continuous control space $\mathcal{A}$.

\textbf{Object Navigation.}
In the traditional object navigation task setting, the robot is given the goal of a target object category $\mathcal{G} = (c_{tgt})$
and is required to navigate to an instance $o_i \in \mathcal{O}$ such that $c(o_i) = c_{tgt}$ as soon as possible, without prior knowledge of the environment layout.

In the more general setting, the robot is given $\mathcal{G} = (c_{tgt}, \Phi)$, where $c_{tgt}$ denotes the target object category, and $\Phi = \{\phi_1, \phi_2, \dots, \phi_K\}$ is a set of semantic constraints.
Each constraint $\phi_k(\cdot)$ may encode object attributes (e.g., color or state) or spatial relationships with other objects, such as \emph{``chair, red''}, \emph{``tv, in the lounge''} or \emph{``person, sitting on a couch''}.
The robot is required to navigate to an object instance $o_i \in \mathcal{O}$ that satisfies $\{c(o_i) = c_{tgt}\} \; \land \; \{\forall \phi_k \in \Phi,\; \phi_k(o_i) = 1\}$, which introduces additional semantic reasoning requirements beyond category-level object navigation.

An episode is considered successful if the robot reaches a pose $p_t$ from which at least one object instance $o_i$ satisfying the task constraints is observable and ${dist}(p_t, o_i) \le \epsilon_{succ}$.

\section{Method}
\label{sec:method}

\label{sec:method_overview}
Our navigation system is decoupled into three levels, each responsible for a distinct aspect of the ObjectNav task: high-level semantic reasoning, mid-level navigation planning, and low-level motion control. Below, we provide detailed descriptions of the design and functionality of each level.





\subsection{High-level - Semantic Reasoning}
\label{sec:method_highlevel}
The high-level semantic reasoning module constructs a structured scene representation of the environment and leverages the reasoning capabilities of the VLM to provide semantic-grounded navigation guidance for the mid-level navigation planning module. It consists of two core components: Scene Representation Building and VLM Reasoning.

\subsubsection{Scene Representation Building}
\label{sec:method_scenerepresentation}
\leavevmode\\
\indent \textbf{Room Node} \raisebox{-0.15\height}{\includegraphics[height=1em]{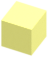}} $v^{r}$:
To decompose the environment into a set of semantically meaningful high-level structural units and exploit this structure for efficient exploration, we introduce room nodes at the top layer of the scene representation, where each room node represents an individual room in the environment.
Inspired by~\cite{werby2024hierarchical,hughes2022hydra}, we identify wall structures in the environment by fitting planar surfaces and analyzing the distribution of point clouds along the vertical ($z$) axis. By dilating the detected wall regions, we partition the initially connected space into multiple disconnected components, each of which is treated as an individual room and added to our representation $\mathcal{R}$ as a room node $v_i^r$.

For each room node $v_i^r$, we maintain a set of attributes
$
\boldsymbol{A}\left(v_i^r\right) = \{ m^r_i,\; c^r_i,\; I^r_i \},
$
where $m_i$ denotes the 2D top-down room mask, $c_i$ represents the room category (e.g., kitchen or bedroom), and $I_i$ is a representative RGB image that provides the best view of the room
\footnote{The best view means the image maximizing the room's visible voxels.}.

\textbf{Viewpoint Node} \raisebox{-0.2\height}{\includegraphics[height=1em]{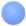}} $v^{v}$:
To efficiently store semantic information and discretely represent explored areas, we introduce viewpoint nodes at the middle layer of the scene representation. Each viewpoint node corresponds to a previously visited location, capturing surrounding semantic and geometric information within a defined range. These nodes together form a compact discrete representation of the environment.

For viewpoint construction, we define a coverage distance $d_{cover}$. Voxels within this distance from a viewpoint are considered fully observed, both semantically and geometrically, forming the coverage region $\mathcal{C}(v_i^v)$ of viewpoint node $v_i^v$.
During exploration, we continuously evaluate the coverage region at the robot's current pose and compare it against the union of coverage regions of all existing viewpoint nodes. Let
$\mathcal{C}_{prev} = \bigcup_{v_i^v \in \mathcal{R}} \mathcal{C}(v_i^v)$
denote the accumulated coverage of viewpoints in representation $\mathcal{R}$.
If the newly observed region provides sufficient novel coverage beyond what has already been captured, i.e.,
$\left| \mathcal{C}_t \setminus \mathcal{C}_{prev} \right| > \epsilon,$
we add the robot's current pose as a new viewpoint node $v_j^v$ to $\mathcal{R}$.

For each viewpoint node $v^{v}_i$, we maintain a set of attributes
$\boldsymbol{A}(v^{v}_i) = \{ p_i, \mathcal{C}_i, I_i \},$
where $p_i$ is the position, $\mathcal{C}_i$ is the coverage region associated with the viewpoint, and $I_i$ is the panorama image captured at the viewpoint.
With this design, RGB images are stored only at representative viewpoint nodes rather than along the entire trajectory, enabling efficient storage of semantic information.

\textbf{Object Node} \raisebox{-0.25\height}{\includegraphics[height=1.15em]{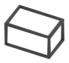}} $v^{o}$:
To model object instances, we introduce object nodes at the lowest layer of the scene representation, where each node corresponds to an object instance in the environment. We use open-vocabulary detection and segmentation methods~\cite{cheng2024yolo,wang2025yoloe,ravi2024sam2} to obtain segmented 3D object instances. At time step $t$, we reconstruct each object's 3D point cloud from the predicted masks, LiDAR data, and robot pose. For each newly detected object, we add an object node $v^{o}_i$ to $\mathcal{R}$ and merge it with existing nodes if they represent the same physical instance.

For each object node $v^{o}_i$, we maintain a set of attributes
$\boldsymbol{A}(v^{o}_i) = \{ c_i, {conf}_i, P_i, {bbox}_i, I_i, (\varphi_1, \varphi_2, ...)\},$
where $c_i$ is the object category, ${conf}_i$ is the detection confidence score, $P_i$ is the reconstructed 3D point cloud of the object, ${bbox}_i$ is the 3D bounding box of the object, $I_i$ is a representative RGB image of the object that provides the best view of the object\footnote{
The best view means the image maximizing the object's viewing angle.}, and $(\varphi_1, \varphi_2, ...)$ are the self-attribute of the object (e.g., color, material). 
To keep the representation compact and general, object self-attributes are inferred on demand rather than predefined.
When a task specifies self-attribute constraints $\phi$ for category $c_i$, we retrieve the corresponding nodes $\{v_j^{o} \: | \: c(v_j^o)=c_i\}$ and use their RGB images to prompt the VLM for attribute inference, appending the results $\varphi$ to each node.
This design efficiently satisfies universal self-attribute constraints while avoiding the redundancy.

\textbf{Room-Room Edge} \raisebox{-0.25\height}{\includegraphics[height=0.9em]{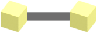}} $e^{r-r}$:
To model room connectivity, we introduce room-room edges. An edge $e^{r-r}_{i,j}$ is added in $\mathcal{R}$ when two room nodes correspond to rooms connected in the environment via a doorway.

\textbf{Room-Viewpoint Edge} \raisebox{-0.25\height}{\includegraphics[height=0.9em]{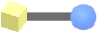}} $e^{r-v}$:
To model viewpoint-room affiliation, we introduce room-viewpoint edges. Edge $e_{j,i}^{r-v}$ is added to $\mathcal{R}$ when $v_i^{v}$ lies in room $v_j^{r}$.

\textbf{Room-Object Edge} \raisebox{-0.25\height}{\includegraphics[height=0.9em]{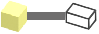}} $e^{r-o}$:
To model object-room containment, we introduce room-object edges. An edge $e_{i,j}^{r-o}$ is added in $\mathcal{R}$ when $v_j^{o}$ lies within room $v_i^{r}$.

\textbf{Viewpoint-Object Edge} \raisebox{-0.25\height}{\includegraphics[height=0.9em]{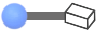}} $e^{v-o}$:
To model object-viewpoint visibility, we introduce viewpoint-object edges. An edge $e_{i,j}^{v-o}$ is added in $\mathcal{R}$ when the robot at viewpoint $v_i^{v}$ can observe object node $v_j^{o}$.

\textbf{Object-Object Edge} \raisebox{-0.25\height}{\includegraphics[height=1.1em]{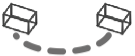}} $e^{o-o}$:
To model spatial relationships between object instances, we introduce object-object edges. To avoid excessive graph density, such edges are added only when required by the task.
Specifically, if the instruction specifies a spatial constraint $\phi$ (e.g., “the cup is on the table”) between $v_i^{o}$ and $v_j^{o}$, we retrieve viewpoint nodes $\mathcal{V}_{i,j} = \{ v_k^{v} \mid e_{k,i}^{v-o} \in \mathcal{R}, e_{k,j}^{v-o} \in \mathcal{R} \}$ that observe both objects.
Using their RGB images, the VLM determines whether $\phi$ holds. If so, we add an edge $e_{i,j}^{o-o}$ in $\mathcal{R}$ with attribute $\boldsymbol{A}(e_{i,j}^{o-o}) = {\varphi}$ indicating the satisfied spatial relation.
This design efficiently satisfies task-specific object-object spatial constraints while avoiding the redundancy.

\subsubsection{VLM Reasoning}
\label{sec:method_vlmreasoning}
In the high-level module, we incorporate a VLM Reasoning component. It leverages the contextual information encoded in the scene representation $\mathcal{R}$ to provide semantic-grounded guidance for navigation through the common-sense reasoning capabilities of VLMs.

\subsection{Mid-level - Room-based Navigation}
\label{sec:method_midlevel}
The mid-level module leverages the structured scene representation and VLM guidance from the high-level module to plan global exploration strategies and navigation paths, outputting waypoints for the low-level control module.
In mid-level, rooms serve as the minimal semantic planning units. Within each room, efficient classical exploration algorithms are used to cover free space, while VLM reasoning is reserved for room-level decision-making.
Accordingly, the module includes two components: an in-room exploration policy for efficient coverage and a cross-room navigation policy that selects and switches rooms under high-level semantic guidance, enabling efficient global object search.


\subsubsection{In-room Exploration Policy}
\label{sec:method_inroom}
Following~\cite{cao2021tare}, our in-room exploration adopts a two-level planning structure consisting of \emph{local planning} and \emph{global planning} to achieve efficient coverage within each room.

When exploring room $v^{r}_j$, we define a local planning horizon 
$\mathcal{H} = \{c_1, \ldots, c_n\} \subset \mathcal{M}$ as a set of sampled pose candidates. 
The traversable candidates within the room are 
$\mathcal{H}^{v^{r}_j}_{{trav}} = \{ c_i \in \mathcal{H} \mid c_i \text{ is traversable } \& \; m^{r}_j(c_i)=1 \}$, 
where $m^{r}_j$ denotes the room mask of $v^{r}_j$.

We define a surface point set $\mathcal{S} \subset \mathcal{M}$ representing the generalized boundary between free and non-free space (including occupied and unknown space). For each pose candidate $c$, let $\mathcal{S}_{{cov}}(c)$ denote the surface points it can cover, which means ${dist}(p, c) < d_{{cover}}$ and $p$ is visible from $c$. Let $\mathcal{T}_{{traj}}$ be the set of previously visited poses, and define the uncovered surface points as
$\hat{\mathcal{S}} = \bigcup_{c \in \mathcal{T}_{{traj}}} \mathcal{S}(c) \setminus \mathcal{S}_{{cov}}.$

For each $c_i \in \mathcal{H}^{v^{r}_j}_{trav}$, the coverage score is 
$w_{cov}(c_i) = \left| \mathcal{S}_{cov}(c_i) \cap \hat{\mathcal{S}} \right|.$
We apply stochastic sampling guided by $w_{{cov}}$ to select pose candidates, iteratively updating the uncovered surface set until all remaining scores fall below $\delta$. 
A TSP is then solved over the selected candidates to generate a local exploration path. 
This procedure is repeated $K$ times, and the minimum-cost path is chosen.

Local and global planning are coordinated through a rolling window mechanism: chosen pose candidates leaving the local horizon are transferred to the global horizon, where a TSP generates the global exploration path. The final path concatenates local and global plans and is converted into waypoints for low-level control.

\subsubsection{Cross-room Navigation Policy}
\label{sec:method_crossroom}


For cross-room navigation, we propose a room-level decision mechanism that uses VLM reasoning to coordinate the exploration order across rooms. The VLM acts as a high-level reasoner, evaluating each room's relevance to the current task online.

\textbf{Early-stop navigation mode.}
When exploring room $v^{r}_i$ and discovering a new room $v^{r}_j$, we provide the VLM with contextual information
$\mathcal{C}_{es} =
\{ \boldsymbol{A}(v^{r}_i),\;
\{ v^o_u \mid e^{r{-}o}_{i,u} \in \mathcal{R} \},\;
\boldsymbol{A}(v^{r}_j),\;
\{ v^o_v \mid e^{r{-}o}_{j,v} \in \mathcal{R} \},\;
\mathcal{G} \}$,
where $\boldsymbol{A}(\cdot)$ and $\{ v^o_\cdot \mid e^{r-o}_{\cdot,\cdot} \in \mathcal{R} \}$ denote room attributes and the object sets belonging to the corresponding room respectively, and $\mathcal{G}$ is the task goal.
Based on $\mathcal{C}_{es}$, the VLM decides whether to terminate exploration in the current room and switch to the new one.
This semantic-aware interruption mechanism enables dynamic focus adjustment and avoids redundant exploration of semantically irrelevant rooms.

\textbf{Room-query navigation mode.}
If in-room exploration in $v^{r}_i$ completes without locating the target, the system switches to a global reassessment mode. 
In this mode, the VLM is provided with contextual information
$\mathcal{C}_{{rq}} =
\{ \{ \boldsymbol{A}(v^{r}_k) \mid v^{r}_k \in \mathcal{R}_{{uncov}} \},\;
{Traj},\;
\mathcal{G} \}$,
including attributes of uncovered rooms, the robot's trajectory and the task goal.
Based on this context, the VLM performs semantic reasoning over the candidate rooms and selects the room that is most likely to contain the target object.
By restricting semantic reasoning to room-level decisions, the system leverages VLM strengths while preserving overall navigation efficiency.

\begin{figure*}[t]
  \centering
  \includegraphics[width=0.99\linewidth]{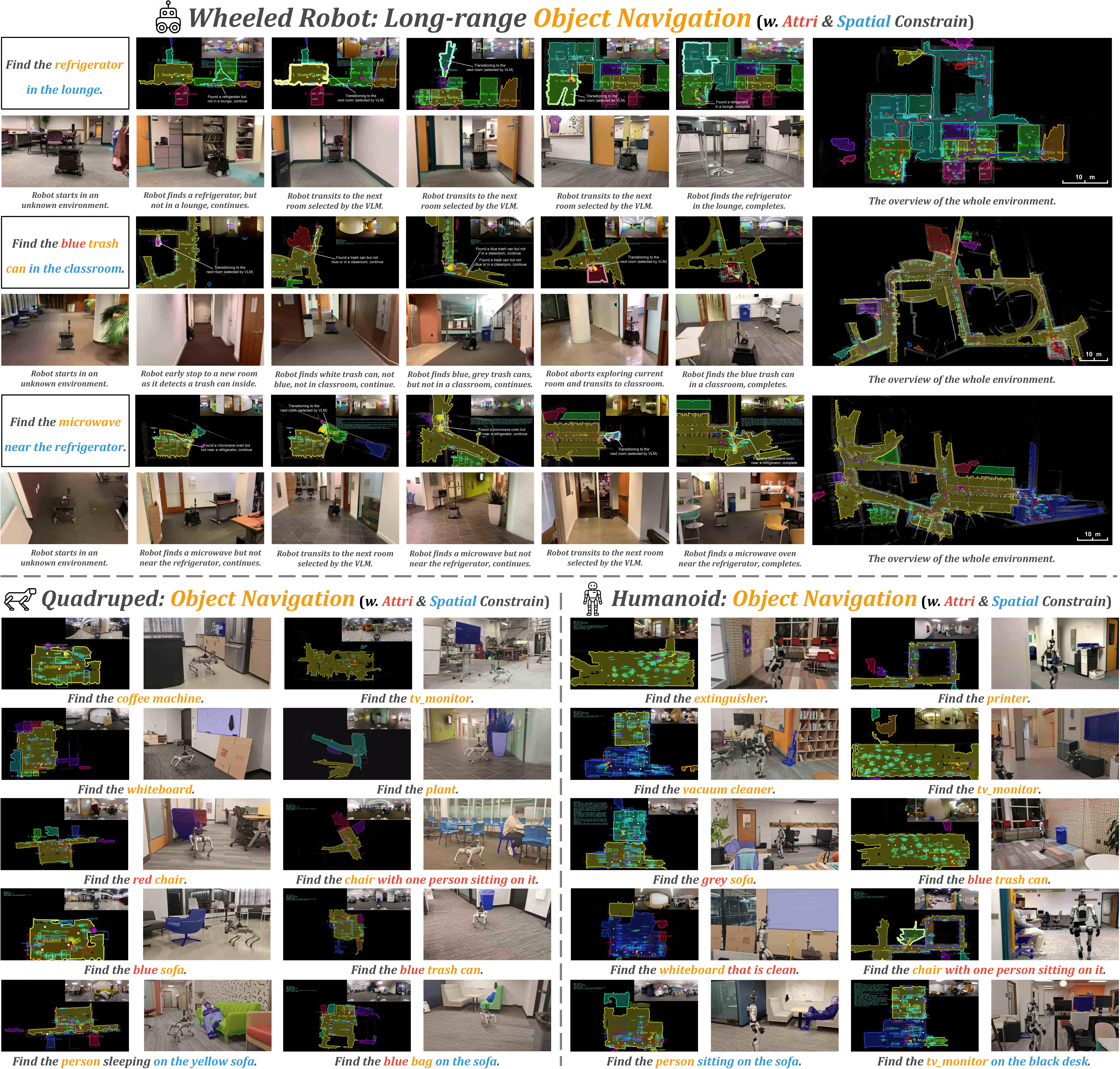}
  \caption{
  \textbf{Qualitative results from real-world deployment of \ourmethodplain.}
  The first three rows show building-scale object navigation with multiple constraints on a wheeled robot, including step-by-step VLM reasoning analysis. 
  The last five rows demonstrate cross-embodiment performance on quadruped and humanoid robots. 
  For each episode, we visualize the final scene representation, first-person view and global overview at task completion (zoom in for details of $\mathcal{R}$).
  \textbf{\textcolor{myred}{Red}} denotes \textbf{\textcolor{myred}{self-attribute constraints}},
  \textbf{\textcolor{myblue}{blue}} denotes \textbf{\textcolor{myblue}{spatial relationship constraints}}
  and \textbf{\textcolor{myorange}{orange}} denotes \textbf{\textcolor{myorange}{target object categories}}.
  }
  \vspace{-15pt}
  \label{fig:real_qual}
\end{figure*}

\subsection{Low-level - Base Autonomy}
\label{sec:method_lowlevel}
To achieve cross-embodiment generality and robust navigation, we design a low-level base autonomy module~\cite{cao2022autonomous} that translates planned waypoints into embodiment-specific motion commands through different motion control modules $M$. It incorporates waypoint-following, collision avoidance and terrain traversability analysis to ensure safe execution.

\begin{table*}[t]
\centering
\caption{
\textbf{Quantitative comparison of \ourmethodplain with state-of-the-art methods on HM3D-v1, HM3D-v2, MP3D, and HM3D-OVON benchmarks.} We report the Success Rate (\textbf{\textit{SR}}) and Success penalized by Path Length (\textbf{\textit{SPL}}) for each method. The best results are highlighted in \textbf{bold}, while the second-best results are {\ul underlined}.
}
\vspace{-3pt}
\label{tab:sim_quant}
\setlength{\tabcolsep}{5pt}
\resizebox{\textwidth}{!}{
\begin{tabular}{lcccccccccc}
\toprule
\multirow{2}{*}{Method}                          & \multirow{2}{*}{Open-Set} & \multirow{2}{*}{Zero-Shot} & \multicolumn{2}{c}{HM3D-v1}                                & \multicolumn{2}{c}{HM3D-v2}                                & \multicolumn{2}{c}{MP3D}                                   & \multicolumn{2}{c}{HM3D-OVON}                              \\ \cmidrule{4-11} 
                                                 &                           &                            & \textbf{\textit{SR}} (\%) $\uparrow$ & \textbf{\textit{SPL}} (\%) $\uparrow$ & \textbf{\textit{SR}} (\%) $\uparrow$ & \textbf{\textit{SPL}} (\%) $\uparrow$ & \textbf{\textit{SR}} (\%) $\uparrow$ & \textbf{\textit{SPL}} (\%) $\uparrow$ & \textbf{\textit{SR}} (\%) $\uparrow$ & \textbf{\textit{SPL}} (\%) $\uparrow$ \\ \midrule
L3MVN~\cite{Yu_2023}                             & \ding{55}                 & \ding{51}                    & 50.4                        & 23.1                         & 36.3                        & 15.7                         & 34.9                        & 14.5                         & -                           & -                            \\
ESC~\cite{zhou2023escexplorationsoftcommonsense} & \ding{51}                 & \ding{51}                    & 39.2                        & 22.3                         & -                           & -                            & 28.7                        & 11.2                         & -                           & -                            \\
VoroNav~\cite{wu2024voronav}                     & \ding{51}                 & \ding{51}                    & 42.0                        & 26.0                         & -                           & -                            & -                           & -                            & -                           & -                            \\
VLFM~\cite{yokoyama2024vlfm}                     & \ding{51}                 & \ding{51}                    & 52.5                        & 30.4                         & 63.6                        & 32.5                         & 36.4                        & 17.5                         & 35.2                        & {\ul 19.6}                    \\
SG-Nav~\cite{yin2024sg}                          & \ding{51}                 & \ding{51}                    & 54.0                        & 24.9                         & 49.6                        & 25.5                         & {\ul 40.2}                  & 16.0                         & -                           & -                            \\
OpenFMNav~\cite{kuang2024openfmnav}              & \ding{51}                 & \ding{51}                    & 54.9                        & 24.4                         & -                           & -                            & 37.2                        & 15.7                         & -                           & -                            \\
TriHelper~\cite{zhang2024trihelper}              & \ding{51}                 & \ding{51}                    & 56.5                        & 25.3                         & -                           & -                            & -                           & -                            & -                           & -                            \\
InstructNav~\cite{longinstructnav}               & \ding{51}                 & \ding{51}                    & -                           & -                            & 58.0                        & 20.9                         & -                           & -                            & -                           & -                            \\
MTU3D~\cite{zhu2025move}                         & \ding{51}                 & \ding{51}                    & -                           & -                            & -                           & -                            & -                           & -                            & {\ul 40.8}                  & 12.1                         \\
ApexNav~\cite{zhang2025apexnav}                  & \ding{51}                 & \ding{51}                    & {\ul 59.6}                  & \textbf{33.0}                & {\ul 76.2}                  & \textbf{38.0}                & 39.2                        & {\ul 17.8}                   & -                           & -                            \\ \midrule
\rowcolor{gray!15} \textbf{\ourmethodplain}      & \ding{51}                 & \ding{51}                    & \textbf{63.7}               & {\ul 30.5}                   & \textbf{80.8}               & {\ul 37.2}                   & \textbf{50.7}               & \textbf{18.1}                & \textbf{54.9}               & \textbf{26.1}                \\ \bottomrule
\end{tabular}
}
\vspace{-13pt}
\end{table*}

\begin{table}[t]
\caption{{Quantitative comparison of \ourmethodplain with the state-of-the-art methods in real-world environments.}}
\vspace{-3pt}
\label{tab:real_quant}
\setlength{\tabcolsep}{2.5pt}
\resizebox{\linewidth}{!}{
\begin{tabular}{l|ccc|ccc|ccc}
\toprule
\multirow{2}{*}{Method} & \multicolumn{3}{c|}{Easy}                                   & \multicolumn{3}{c|}{Medium}                               & \multicolumn{3}{c}{Hard}                                 \\ \cmidrule(lr){2-4} \cmidrule(lr){5-7} \cmidrule(lr){8-10}
                        & \textbf{\textit{SR}}$\uparrow$   & \textbf{\textit{SPT}}$\uparrow$  & \textbf{{\textit{AT}}}$\downarrow$ & \textbf{\textit{SR}}$\uparrow$   & \textbf{\textit{SPT}}$\uparrow$  & \textbf{\textit{AT}}$\downarrow$ & \textbf{\textit{SR}}$\uparrow$   & \textbf{\textit{SPT}}$\uparrow$  & \textbf{\textit{AT}}$\downarrow$ \\ \midrule
VLFM~\cite{yokoyama2024vlfm}                    & 58.3           & 40.3          & 52.1                   & 47.5          & 34.4          & 75.3                  & 25.6          & 12.9          & 97.4                  \\
Instruct~\cite{longinstructnav}                & 45.8           & 29.5          & 67.4                   & 55.0          & 27.8          & 84.6                  & 37.2          & 20.7          & 103.4                 \\
\ourmethodplain        & \textbf{100.0} & \textbf{83.8} & \textbf{29.1}          & \textbf{97.5} & \textbf{67.9} & \textbf{72.8}         & \textbf{98.3} & \textbf{71.8} & \textbf{67.6} \\
\bottomrule
\end{tabular}
}
\vspace{-17pt}
\end{table}

\section{Experiment}
\label{sec:experiment}
\subsection{Experiment Setup}
\label{sec:experiment_setup}
We evaluate \ourmethodplain in both real-world environments and simulation benchmarks. In real-world, we conduct 190 episodes, including 78 episodes for qualitative evaluation and 112 episodes for quantitative evaluation. In simulation, we conduct 8,195 episodes across four different benchmarks.

\textbf{Real-World Robot Setup:}
In real-world experiments, we deploy our system on three different robot platforms: a Mecanum wheeled robot~\cite{autonomy_stack_mecanum_wheel}, Unitree Go2 quadruped and Unitree G1 humanoid. For each platform, we mount a Livox Mid-360 LiDAR and a Ricoh Theta Z1 panoramic camera for sensor inputs. For Unitree Go2 and G1, the built-in LiDAR is additionally used for locomotion control. \ourmethod runs on a laptop with an i9-14900HX CPU and an RTX-4090 GPU.

\textbf{Simulation Benchmark Setup:}
In simulation, we follow the Habitat Challenge 2023~\cite{habitatchallenge2023} setup. The robot is equipped with a 640$\times$480 RGB-D camera.
The agent moves forward by 0.25m per step and rotates by $30^\circ$ per action. We convert depth images into point clouds using the robot pose as input to our system. We conduct experiments on four benchmarks: (1) HM3D-v1~\cite{habitatchallenge2022}, including 2,000 episodes across 20 high-fidelity scenes with 6 target categories; (2) HM3D-v2~\cite{puig2023habitat3}, including 1,000 episodes across 36 high-fidelity scenes with 6 target categories; (3) MP3D~\cite{Matterport3D}, including 2,195 episodes across 11 scenes with 21 target categories; and (4) HM3D-OVON~\cite{yokoyama2024hm3d} val-unseen split, including 3,000 episodes across 36 scenes with 49 target categories.

\textbf{Evaluation Metrics:}
Following~\cite{habitatchallenge2023}, we use two primary metrics: (1) Success Rate (SR), defined as the percentage of episodes in which the robot reaches the target object within a predefined success distance; and (2) Success weighted by Path Length (SPL), which penalizes the success rate by the ratio of the shortest path length to the actual path length. In addition, for real-world experiments where ground-truth shortest path lengths are unavailable, we adopt Success Penalized by Time (SPT) as an alternative metric, defined as $SPT = SR \cdot (1 - \frac{t}{T_{timeout}})$, where $t$ is the task completion time and $T_{timeout}$ is the predefined timeout threshold. We also report the Average Time (AT) over successful episodes.

\textbf{Implementation Details:}
For semantic mapping, we use YOLOv8x\-worldv2~\cite{cheng2024yolo} and SAM2.1\_hiera\_b+~\cite{ravi2024sam2} for object detection and segmentation. For the VLM module, we use Gemini-2.5-flash~\cite{team2023gemini}.
{For the Unitree Go2 and G1, we use the built-in locomotion policy for low-level control.}

\subsection{Qualitative Results in Real-world}
\label{sec:qualitative_results_real_world}
We conduct 78 real-world episodes for qualitative evaluation, including 4 in building-scale environments, 41 using a wheeled robot platform, 19 using a quadruped robot, and 14 using a humanoid robot. The results are shown in \cref{fig:real_qual}. The first three rows illustrate performance in building-scale environments on the wheeled robot platform.
The results demonstrate that the semantic mapping module successfully models complex environments into a structured scene representation, which serves as a unified representation for high-level reasoning and decision-making.
Based on this representation, the VLM Query module performs instruction-aware reasoning over semantic constraints and provides semantically grounded navigation guidance to the planning module. For example, in the second row, the high-level module identifies the classroom in the environment and guides the robot to prioritize exploring that room, thereby improving search efficiency.
Additionally, the VLM Query module determines whether objects satisfy semantic and spatial constraints based on the structured scene representation. As shown in the third row, the system identifies a microwave and a refrigerator and further evaluates whether their spatial relationship satisfies the semantic constraint “near,” enabling accurate localization of the target object.
The last five rows show results on quadruped and humanoid platforms, demonstrating strong cross-platform generalization and robustness under diverse semantic constraints.
Additional results and videos are available on our project page: \href{https://cmu-vln.github.io/}{\textit{cmu-vln.github.io}}.

\subsection{Quantitative Results in Real-world}
\label{sec:quantitative_results_real_world}
For quantitative evaluation, we conduct 112 episodes on the wheeled robot platform across 10 target object categories and 3 real-world environments. We categorize the environments into easy, medium, and hard difficulty levels.
The easy setting (24 episodes) contains a single room with the target mostly in the initial view. The medium setting (40 episodes) also contains a single room but requires moderate exploration. The hard setting (48 episodes) consists of three rooms, where the target is located in a different room from the robot's initial position, requiring cross-room exploration. Timeouts are 3 minutes for easy and 4 minutes for medium and hard.
For fair comparison, all baselines share the same point cloud and panoramic inputs as well as the same low-level base autonomy stack.
We report SR, SPT, and AT (\cref{tab:real_quant}), with AT computed over successful episodes. Our system significantly outperforms the baseline across all difficulty levels. In the hard setting, SR and SPT improve by 61.1\% and 51.1\%, while AT is reduced by 29.8 seconds.
These results demonstrate that with the proposed semantic reasoning and room-based navigation strategy, our system effectively leverages structural and semantic information for navigation, substantially improving success rate and efficiency. 
The slightly better performance in the hard setting is because multi-room layouts introduce limited additional difficulty for our system, however the medium scene contains denser obstacles that slow the robot, reducing SPT and AT.

\subsection{Quantitative Results in Simulation}
\label{sec:quantitative_results_simulation}
We compare our system with multiple state-of-the-art baselines on four simulation benchmarks, with results shown in \cref{tab:sim_quant}. Overall, our system consistently outperforms the baselines across all benchmarks. In particular, on the HM3D-OVON benchmark, SR and SPL improve 14.1\% and 6.5\% respectively. These results demonstrate that our approach effectively exploits structural and semantic information for navigation, significantly improving task efficiency and success rate.
The smaller improvement in SPL compared to SR arises from our real-world-oriented design, which adopts stricter coverage strategies to handle environmental complexity and sensor noise. This can cause slight over-coverage in simulation, resulting in more modest SPL improvements.
\section{Conclusion}
\label{sec:conclusion}
In this work, we present a three-level ObjectNav system designed for real-world deployment and cross-embodiment generalization. By decoupling semantic reasoning, navigation planning, and motion control, our system enables each component to operate at the appropriate level, balancing semantic intelligence with spatial reliability. A structured three-layer scene representation supports interpretable and grounded VLM reasoning, while the hierarchical room-based navigation strategy integrates global semantic decision-making with efficient local exploration.
Extensive real-world experiments across three robot embodiments demonstrate that our system achieves substantial improvements in both success rate and navigation efficiency. To the best of our knowledge, this work is the first to reliably and efficiently accomplish building-scale long-range object navigation in complex real-world environments. Results on multiple simulation benchmarks further validate its robustness and generality.

\vspace{-3pt}






\section*{ACKNOWLEDGMENT}
This work is partially sponsored by Denso and Alpha-Z. The humanoid robot hardware is provided by Dimensional.


\bibliographystyle{IEEEtran}
\bibliography{IEEEabrv,main}

\end{document}